# *Seeing Voices and Hearing Faces:* Cross-modal biometric matching

Arsha Nagrani    Samuel Albanie    Andrew Zisserman

VGG, Dept. of Engineering Science, University of Oxford

{arsha, albanie, az}@robots.ox.ac.uk

## Abstract

*We introduce a seemingly impossible task: given only an audio clip of someone speaking, decide which of two face images is the speaker. In this paper we study this, and a number of related cross-modal tasks, aimed at answering the question: how much can we infer from the voice about the face and vice versa?*

*We study this task "in the wild", employing the datasets that are now publicly available for face recognition from static images (VGGFace) and speaker identification from audio (VoxCeleb). These provide training and testing scenarios for both static and dynamic testing of cross-modal matching. We make the following contributions: (i) we introduce CNN architectures for both binary and multi-way cross-modal face and audio matching; (ii) we compare dynamic testing (where video information is available, but the audio is not from the same video) with static testing (where only a single still image is available); and (iii) we use human testing as a baseline to calibrate the difficulty of the task. We show that a CNN can indeed be trained to solve this task in both the static and dynamic scenarios, and is even well above chance on 10-way classification of the face given the voice. The CNN matches human performance on easy examples (e.g. different gender across faces) but exceeds human performance on more challenging examples (e.g. faces with the same gender, age and nationality)[1].*

## 1. Introduction

Can you recognise someone's face if you have only heard their voice? Or recognise their voice if you have only seen their face? As humans, we may *'see voices'* or *'hear faces'* by forming mental pictures of what a person looks like after only hearing their voice, or vice versa. This phenomenon has been investigated in a number of studies on human perception and neurology [19, 44], where participants completed a sequential two-alternative forced choice matching task. They were asked to listen to a human voice (Voice X), and then pick the face corresponding to the same identity, between two still/dynamic face images. It is also a familiar trope in Hollywood films that someone can be recognised after only hearing their voice – for example in the film "Die Hard", where the Bruce Willis character (John Mclane) emerges from the building towards the end of the film and is instantly able to recognise the cop (Sgt. Al Powell) who he has only spoken to by radio throughout the film, but never seen. This task of *cross-modal* (face and voice) recognition, or 'cross-modal biometric matching', is the objective of this paper. As illustrated in figure 1, there are two related tasks: first, given an image or video of a face, determine which of two or more voices it corresponds to; second, and conversely, given an audio clip of a voice, determine which of two or more face images or videos it corresponds to. Note, the voice and face video are not acquired simultaneously, so methods of active speaker detection that may rely on synchronisation of the audio and lip motion, e.g. [11] cannot be employed here.

---

[1]Data, models and appendices can be found at http://www.robots.ox.ac.uk/~vgg/research/CMBiometrics

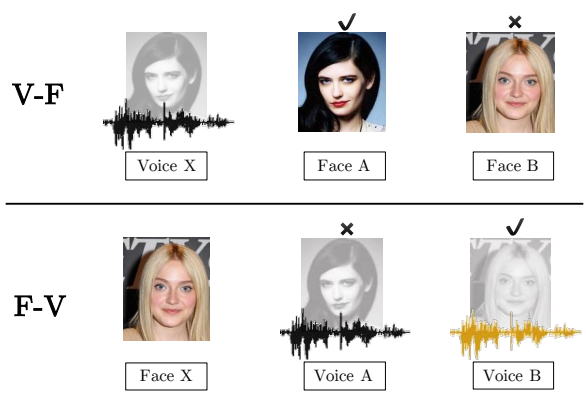

Figure 1: We introduce the task of cross-modal biometric matching, and consider two specific formulations of the problem: *(Top) V-F*: given an audio clip of a voice and two or more face images/videos, select the face image/video that corresponds to the voice. *(Bottom) F-V*: given an image or video of a face, determine the corresponding voice.

That this task might be possible at all is due to the existence of factors that are common to both modalities; in particular, specific latent properties (like age, gender, ethnic-ity/accent) influence both the facial appearance and voice. Besides these, there exist other, more subtle cross-modal biometrics. Studies in biology and evolutionary perception [52] show that hormone levels during puberty affect both face morphology and voice pitch. In males, higher testosterone-oestrogen ratios lead to a prominent eyebrow ridge, broad chin, small eyes, and thin lips [48], while vocal folds situated in the larynx also increase in size, thus leading to a lower voice pitch [17]. Similarly for females, higher oestrogen levels cause large eyes and full lips [48] and prevent the vocal folds from enlargement, leading to higher voice pitch [17]. Besides the above static properties, given a video stream, we expect there to exist more (dynamic) cross-modal biometrics. For example, the 'manner of speaking' can be an important cross-modal biometric. Sheffert and Olson [40] suggested that visual information about a person's particular idiosyncratic speaking style is related to the speaker's auditory attributes. The origins of this link lie in the mechanics of speech production, which, when shaping the vocal tract, determines both facial motion as well as the sound of the voice [19].

Apart from establishing that it is indeed possible to solve cross-modal biometric matching, which is an interesting scientific result on its own, there are also practical applications of the technology – not least in surveillance. Imagine the following scenario: the only information we have about a person is a handful of speaking (audio) samples, because the data was recorded from telephone conversations. We then want to identify the individual from a video stream, for example from CCTV. A more benign application is automatically labelling characters in TV and film material where characters may be heard but not seen at the same time, and so cross-modal matching can be used to infer the labels.

In this paper we approach the problem using the tools of deep learning trained on large-scale datasets. We make the following contributions: first, we introduce a CNN architecture that ingests face images and voice spectrograms, and is able to infer the correspondence between them. The network is trained on a large-scale dataset of voices (Vox-Celeb [33]) and faces (VGGFace [36]) from the same identities. Second, we investigate the performance of the network using still, dynamic images, or both. We show, in contrast to the findings in the perception literature, that the task can be solved far better than chance using static images alone, and that the performance improves further using dynamic images. We also carry out our own study of human performance using AMT. Finally, we generalise the two-alternative forced choice architecture to multi-way classification and report results for this more challenging task.

## 2. Related Work

**Human Perception Studies:** The broad consensus among studies exploring cross-modal matching of faces and voices using human participants, is that matching is only possible when dynamic visual information about articulatory patterns is available [19, 26, 37]. In particular, works have demonstrated coupling between an individual's idiosyncratic speaking style, the sound of their voice and the manner in which their face moves [13, 27, 53], suggesting the presence of dynamic information which can be exploited to solve the matching task. Although these studies demonstrate that static face voice matching performance lies at chance level [19, 26], we note that there has been research which challenges this perspective [24, 43]. However, while Krauss et al. [24] showed that people could match a voice to one of two static images with above-random accuracy, the stimuli were *full-body* images rather than images of faces, which may have provided additional cues to inform accurate matching (see [44] for a detailed discussion of these contradictory results). It is worth noting that the difficulty of the task is highly dependent on the specific stimuli sets provided—as we show in this work, some face-voice combinations are more distinctive than others.

**Face Recognition and Speaker Identification:** The tasks of face recognition and speaker identification are longstanding problems in the vision and speech research communities, and consequently an in-depth review of these topics is beyond the scope of this work. However, we note that the recent advent of deep CNNs with large datasets has considerably advanced the state-of-the-art in both face recognition [21, 36, 46, 47] and speaker recognition [14, 33, 39, 45]. Unfortunately, while these recognition models have proven remarkably effective at representation learning from a single modality, the alignment of learned representations across the modalities is less developed. In this work we address this issue through the development of a multimodal architecture that directly ingests data from both faces and voices and learns a correspondence between them.

**Cross-modal Matching:** Cross-modal matching has received considerable attention using visual data and text (natural language). Methods have been developed to establish mappings from images [16, 20, 23, 25, 50] and videos [49] to textual descriptions (e.g. captioning), generating visual models from text [51, 57] and solving visual question answering problems [1, 29, 31]. In cross-modal matching between video and *audio* however, work is limited, particularly in the field of biometrics (person or speaker recognition). Recent work has begun to explore the tasks of audio-visual matching for scenes and objects [2, 3, 4, 35] and audio-visual speech recognition (lip reading [12], lip sync [11] etc). In biometrics, there has also been work that uses both modalities to improve performance [7, 22] but not one to recognise the other. Le and Odobez [28] use transfer learning from face embeddings to try and improve speaker diarisation results. The only attempt we can find to solve a similar task to the one proposed here (but only for videos,

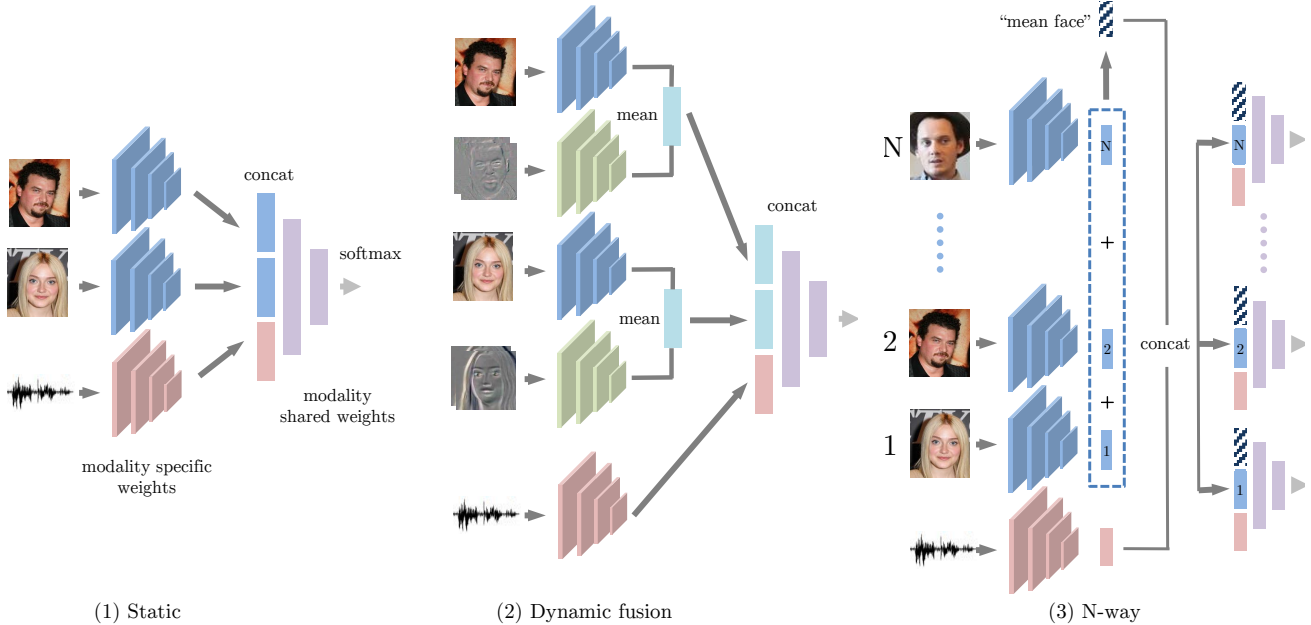

(1) Static     (2) Dynamic fusion     (3) N-way

Figure 2: The three main networks architectures used in this paper. From left to right: (1) The *static* 3-stream CNN architecture consisting of two face sub-networks and one voice network, (2) a 5-stream *dynamic-fusion* architecture with two extra streams as dynamic feature sub-networks, and finally (3) the *N-way classification* architecture which can deal with any number of face inputs at test time due to the concept of query pooling (see Sec. 3.2). Voice-only weights shown in red, face-only weights in blue/green (static/dynamic), and modality-shared weights in purple (best viewed in colour). Modality-specific weights of the same colour are shared amongst different inputs.

and not still face images) is by [38]. This work seeks to map a statistical model of the features in one modality to a statistical model of the features in another modality. It is evaluated on the M2VTS audio-visual database for 25 male subjects who count from zero to nine. In contrast, we aim to solve this task at large-scale, 'in the wild', and with longer more natural speech segments from unconstrained interview videos.

## 3. Cross-Modal Models

For the task of forced matching between two faces and voice input (V-F formulation), our objective is to identify which of a pair of given faces possesses the same identity as the voice. Since this problem admits a natural symmetry with the F-V formulation (matching between two voices and a face), each component of our method can be readily adapted to address either task. For notational clarity, we focus our description on the V-F formulation. The forced matching task can be defined as follows; let $x = \{v, f_1, f_2\}$ denote a set consisting of an anchor voice segment $v$ and two face images $f_1$ and $f_2$. Each input set $x$ contains one positive and one negative face, where face $f_i$ is defined as positive if it possesses the same identity as the anchor voice, and negative otherwise. We pose the matching task as a binary classification problem, in which the objective is to predict the position $y \in \{1, 2\}$ of the positive face. Given images and voices of known identity, we can construct a dataset of training examples $\mathcal{D} = \{x_n, y_n\}_{n=1}^{N}$ by simply randomising the position of the positive face in each face pair. The learning problem corresponds to maximising likelihood: $\theta = \operatorname{argmax}_\theta L(g_\theta; \mathcal{D})$, where $g_\theta$ is the parameterised model to be learned. The loss to be minimised can then be framed as a cross-entropy loss on target label positions.

We instantiate $g_\theta$ as a three-stream convolutional neural network, taking inspiration from the *odd-one-out* network architecture proposed in [15]. Our forced matching task, however, is unique in the sense that we would like to perform it across two different *modalities*. Our model design consists of three modality specific sub-networks (or streams); two parameter-sharing face sub-networks that ingest image data and a voice sub-network which ingests spectrograms. The three streams are then combined through a fusion layer (via feature concatenation) and fed into modality-shared fully connected layers on top. The fusion layer is required to enable the network to establish a correspondence between faces and voices.

Our model hence has two kinds of layers, modality specific (face and voice) layers and higher-level layers which are shared between both modalities. Similar to the motivation in [3], the rationale behind this architecture is to force early layers to specialise to modality specific features (such as edges in face images and spectral patterns in audio segments), while allowing later layers to capture higher-level latent cross-modal variables (such as gender, age, ethnicity and identity). For the sake of clarity, we state here the three main tasks that we solve in this paper: 1) Static matching, which uses only still face images, 2) Dynamic matching, which involves videos of faces during speech, and 3) N-way classification, which is an extension of the matching task to any number of faces (greater than two). These tasks are described in more detail in section 5. In order to capture dynamic facial appearance, we introduce an additional sub-network which ingests dynamic features extracted from videos. To motivate the design of each of these sub-networks, we next discuss the input representations upon which they will operate.

### 3.1. Input Representations

**Voices:** The input to the voice stream is a short term magnitude spectrogram extracted directly from raw audio. The audio stream is extracted from the video and converted to single-channel, 16-bit streams at a 16kHz sampling rate for consistency. Spectrograms are then generated in a similar manner to that in [33], giving spectrograms of size $512 \times 300$ for three seconds of speech. We perform mean and variance normalisation on every frequency bin of the spectrum, but apply no further speech-specific preprocessing (e.g. silence removal, voice activity detection, or background noise suppression).

**Static Faces:** Each input to the face stream consists of an RGB image, which has been cropped from a source image to contain only the region of the image surrounding a face. The locations of these crops are provided by the datasets used in our experiments (discussed further in Sec. 4)[2]. The resulting region is then resized to a fixed $224 \times 224$ input.

**Dynamic Faces:** The annotated face regions contained in video data are processed as *face-tracks*, defined to be contiguous sequences of frames possessing the same identity. To exploit dynamic cross-modal information from idiosyncratic speaking styles, an estimate of motion is required. Previous work that seeks to perform speaker recognition solely with visual information [8, 9, 34, 56] tends to focus primarily on the lip region. While useful biometric information is concentrated around the lips, (e.g. when uttering the same phoneme or word, different speakers have different mouth shapes [34]), we hypothesise that the motion of other facial features, e.g. eyes or eyebrows, or even the entire motion of the head during speech, could be useful bio-metric cues for identification. We would therefore like to work with a representation of this data that is capable of extracting temporal information from each full face-track.

A wide range of approaches have been proposed to enable CNNs to exploit temporal information from video, including 3D convolutions [18], optical flow [41] and dynamic images [6] which have proven to be particularly effective in the context of human action recognition. In this work, we employ the dynamic image representation, which computes a fixed size representation of a video sequence by learning a ranking machine on the raw pixel input across a given sequence of frames. See section 7 for variant implementation details.

### 3.2. Architectures

**(1) Static Architecture:** Our base architecture comprises two face sub-networks and one voice sub-network. Both the face and voice streams use the VGG-M architecture [10], which achieves a good trade-off between efficiency and performance. The features from each stream are fused through concatenation to form a 3072-dimensional feature[3], which is then processed by three fully connected layers with hidden units of dimensionality 1024, 512 and 2 respectively. Further details of each sub-network can be found in the appendix.

**(2) Dynamic-Fusion Architecture**: Motivated by the effectiveness of dual stream architectures that combine RGB images with temporal features in action recognition [6, 41], we also explore a variant of the base architecture which includes an additional dynamic image stream for each input face. The features computed for each face (RGB + dynamic) are combined after the final fully connected layer in each stream through summation. In more detail, given an augmented input set $x = \{v, f_1, f_2, d_1, d_2\}$, where $d_1$ and $d_2$ are dynamic face inputs, we compute the representation

$$\phi(x) = \phi_2 \circ \Big[(\phi_f(f_1)+\phi_d(d_1))\oplus(\phi_f(f_2)+\phi_d(d_2))\oplus\phi_a(v)\Big]$$

where $\oplus$ denotes concatenation, $\phi_f$ represents the RGB face sub-network, $\phi_d$ the dynamic image face sub-network, $\phi_a$ the operations of the audio sub-network, and $\phi_2$ the modality-shared fully connected layers on top. The two static face streams and two dynamic face streams share separate weights, allowing the different types of face input to be treated accordingly.

**(3) N-way Classification Architecture:** We further extend the architecture to deal with the more challenging task of developing a general cross-modal biometric system that is capable of solving an $N:1$ identification problem. The input to this network consists of an anchor voice segment $v$, 1 positive face and $N-1$ negative faces. As before, the target label $y \in \{1, 2, \ldots, N\}$ denotes the position of the positive face, resulting in an $N$-way classification problem.

---

[2]Since in both datasets, the specified face regions yield a tight face crop, we expand all crops by a factor of $\times 1.6$ to incorporate additional context into the face region.

[3]Each of the three sub-networks produces a 1024-dimensional vector.

|  | Train | Val | Test | Total |
|---|---|---|---|---|
| # of identities | 942 | 116 | 189 | 1,247 |
| *VGGFace Dataset* | | | | |
| # of face images | 873,382 | 47,759 | 74,564 | 995,705 |
| *VoxCeleb Dataset* | | | | |
| # of speech segments | 116,480 | 14,630 | 22,376 | 153,486 |
| # of videos | 16,820 | 2,044 | 3,425 | 22,295 |

Table 1: Statistics for the VGGFace and VoxCeleb datasets. The numbers shown here are only for the overlapping identities in the two datasets.

As a consequence of using concatenation as a fusion layer in our base architecture, the number of face streams cannot be adjusted during inference. This shortcoming is common to many CNN architectures, where it is difficult to change the number of inputs at test time. One approach to resolving this issue would be to concatenate the voice to each face stream separately, however in this scenario each face stream would be unaware of the presence of the other streams. In order to avoid this problem, we add a mean pooling layer to each face stream which calculates the 'mean face' of all the faces in a particular query, thereby making each stream *context aware*. We refer to this simple concept as 'query pooling'.

## 4. Datasets and Training

Due to the novel nature of the task explored in this work, no large-scale public benchmarks exist for evaluating our approach. We therefore construct a new dataset to train and evaluate our method by combining two available datasets with overlapping identities:

**VGGFace [36]:** VGGFace is a large-scale dataset of still face images collected from search engines. We use the 'curated' version of this dataset.

**VoxCeleb [33]:** VoxCeleb is a large-scale audio-visual dataset of human speech collected from YouTube videos. Since this dataset is collected 'in the wild', it covers a wide range of different recording environments and background noise levels. This dataset contains both video and audio.

For the purposes of this task, we use only the data for the $1,247$ identities that overlap between the two datasets.

**Train/Test Split:** Identities in the train and test datasets do not overlap. All speakers whose names start with 'A' or 'B' are reserved for validation, while speakers with names starting with 'C', 'D', 'E' are reserved for testing. This yields a good balance of male and female speakers (dataset statistics are given in table 1).

**Gender, Nationality and Age (GNA) Variation**: To enable a more thorough analysis of our method, gender and nationality labels for speakers in the dataset were obtained by crawling Wikipedia. Note that we crawl for *nationality*, and not *ethnicity*, since this is a variable typically more informative of accent. The distribution of nationalities across the dataset is given in Appendix B. We use these labels to construct a more challenging test set, wherein each triplet contains speakers of the same gender, broad age bracket (speakers between the ages of 30-50 years old were selected manually), and nationality (we restrict to U.S nationals). We note that these conditions are similar to those established during the human perception studies discussed previously [19, 26, 37].

In the sections that follow, we refer to each input $x$ containing two face (either from stills or video) and one voice representation as a *triplet*.

### 4.1. Training Protocol

All networks are trained end-to-end using stochastic gradient descent with batch normalisation. We use a minibatch size of $64$, momentum ($0.9$), weight decay ($5E-4$) and a logarithmically decaying learning rate (initialised to $10^{-2}$ and decaying to $10^{-8}$). The face and voice sub-networks are initialised using the pre-trained weights from the VGGFace and VoxCeleb models trained for face and speaker identification respectively, while the modality shared weights are initialised from a Gaussian distribution. When processing face images, we apply the data augmentation techniques used on the ImageNet classification task by [42] (i.e. random cropping, flipping, colour shift). For the audio segments, we change the speed of each segment by choosing a random speed ratio between 0.95 to 1.05. We then extract a random 3s segment from the audio clip at train time. Training uses $1.2M$ triplets that are selected at random (and the choice is then fixed). Networks are trained for 10 epochs, or until validation error stops decreasing, whichever is sooner.

## 5. Experiments

### 5.1. Tasks

**Static Matching:** Under the static evaluation, each test sample consists of two *static* face images and single speech segment. To construct the test set for this benchmark, we use audio segments from VoxCeleb [33] and face images from VGGFace [36]. We make use of both still images from VGGFace and frames extracted from the videos in the VoxCeleb dataset during training. When processing frames extracted from the VoxCeleb videos, we ensure that the audio segments and frames in a single triplet are not sourced from the same video.

**Dynamic Matching:** The dynamic evaluation assesses performance on videos of human speech. In addition to static cross-modal biometrics, in this setting a person's 'manner of speaking' may also provide important source of identity information. The dataset for this benchmark consists of videos and audio both extracted from VoxCeleb [33]. A triplet in this case consists of two face-tracks and one audio segment.

For the purposes of this task, it is important to minimise any correlation or mutual information based on audio-visual synchrony which could arise if the audio and visual data were extracted at the same time (for example: mouth motion based on the exact lexical content of the sentence, the emotional state of the speaker etc.). While interesting in their own right, these factors do not constitute cross modal biometrics for person verification, and therefore exploiting their presence to solve the matching task is not the objective of this work; we wish to be sensitive *only* to identity. Hence we ensure that the audio segments and face-tracks in a single triplet are not sourced from the same video. While we experiment with different methods for extracting dynamic information from a face-track (described in detail in section 7), the best results were obtained using dense sampling in order to obtain multiple aligned RGB and dynamic images from each face-track. These inputs are then fed into the fusion architecture.

**N-way Classification:** We also extend the V-F task to one of $1:N$ classification. It is important to note that such a task is extremely challenging, particularly since as $N$ increases, the likelihood of solving the problem using isolated variables such as age, gender or ethnicity in isolation (or in combination) diminishes. We use the N-way classification architecture (figure 2, right) to tackle this task. This architecture allows us to train with any number of face images $N_{Tr}$, and then test with any number of test images $N_{Te}$, where $N_{Tr}$ does not have to be equal to $N_{Te}$. We experimented with different values of $N_{Tr} = 2, 3, 5$, however we found that changing the number of faces at train time did not significantly improve results. We therefore report results of accuracy $A_I$ vs $N_{Te}$ trained using $N_{Tr} = 2$ (figure 3).

**Evaluation Protocol:** The static and dynamic cases are evaluated on $10,000$ triplets randomly chosen from the test set. This gives a good balance of easy and difficult triplets. The N-way case is evaluated on $10,000$ inputs, again chosen randomly. Since RGB frames and dynamic images are extracted densely for the dynamic case, at test time we adopt a simple ensembling strategy. Frame predictions are averaged to give a single prediction per triplet. Since we may have two face-tracks of differing lengths in each triplet, the frames from the longer face-track are selected using a stride $s$, where $s = \left\lfloor \frac{length(F_1)}{length(F_2)-1} \right\rfloor$ and $F_1$ and $F_2$ are the two face-tracks. For all three of the above cases, we use the entire audio segment at test time with standard average pooling, following the exact procedure used in [33].

### 5.2. AMT Human Baselines:

Since there are no prior baselines to compare to, it is useful to have a measure of how well humans are able to perform forced matching between faces and voices. Direct comparison with studies on human perception [19, 26, 37] is infeasible given the likely difference in distributions of datasets. We therefore calibrate the difficulty of our dataset by performing our own human study on Amazon Mechanical Turk (AMT). For this study, a set of 500 triplets were randomly sampled from the static test set. Each sample was shown to 20 different workers on AMT in batches of five triplets. An in-depth description of the study can be found in appendix C, and the results are shown in table 2.

### 5.3. Evaluation Measures

We define two metrics to evaluate performance; *Identification Accuracy* and *Marginal Accuracy*. Following the notation introduced in Sec. 3, let $D = \{x_n, y_n\}_{n=1}^{N}$ denote a set of labelled examples where each input triplet takes the form $x_n^{(i,j)} = \{v^{(i)}, f^{(i)}, f^{(j)}\}, i, j \in \mathcal{I}$ (here $\mathcal{I}$ denotes the set of identities). We define the identification accuracy of a predictive model $g$ as follows:

$$A_I = \frac{1}{N} \sum_n |g(x_n^{(i,j)}) = y_n|$$

We further define the marginal accuracy of a predictive model $g$ as:

$$m_A(s) = \frac{1}{\mathcal{N}_s} \sum_{(i=s \vee j=s)} |g(x_n^{(i,j)}) = y_n|$$

where $\mathcal{N}_s := |\{x_n^{(i,j)} : i = s \vee j = s\}|$ represents the number of triplets containing the speaker $s$. Identification accuracy provides a measure of performance on the entire test set, while marginal accuracy enables us to determine speaker-specific performance.

## 6. Results and Discussion

**Static and Dynamic Matching:** We report the results of both the F-V and the V-F formulations for the static and dynamic cases in table 2. The results for the dynamic task are better than those for the static task (by more than 3% for the V-F case). Since the identities in the two datasets are exactly the same, we infer that this increase in accuracy may be due to the presence of visually dynamic information from articulatory patterns.

**N-way classification:** The accuracy $A_I$ vs $N_{Te}$ for a model trained using $N_{Tr} = 2$ is shown in figure 3. As observed from figure 3, although $A_I$ reduces as the number of faces at test time $N_{Te}$ increases, the ratio $A_I/A_R$ indicating the relative improvement of the proposed system compared to chance $A_R$ remains relatively stable, validating the efficacy of our approach.

**V-F vs F-V cases:** The similar accuracies suggest that the task is highly symmetric in nature, aligning closely with the outcome of a prior study in human perception [19].

|  | Static Test | | Dynamic Test | |
| --- | --- | --- | --- | --- |
|  | $A_I$(Total) | $A_I$(GNA-var removed) | $A_I$(Total) | $A_I$(GNA-var removed) |
| V-F | 81.0 | 63.9 | 84.3 | 67.4 |
| F-V | 79.5 | 63.4 | 82.9 | 65.6 |
| Human Baseline (V-F) | 81.3 | 57.1 | - | - |

Table 2: Results are reported using % Identification Accuracy $A_I$ which is calculated using 10,000 test triplets. Since this is a 2-way forced matching task, chance is 50%.

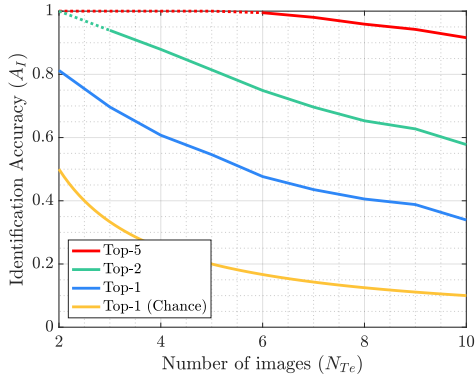

Figure 3: Top-1, top-2 and top-5 identif. acc. ($A_I$) vs the number of face images $N_{Te}$ at test time. As can be seen from the graph, the model performs well above chance for all values of $N_{Te}$.

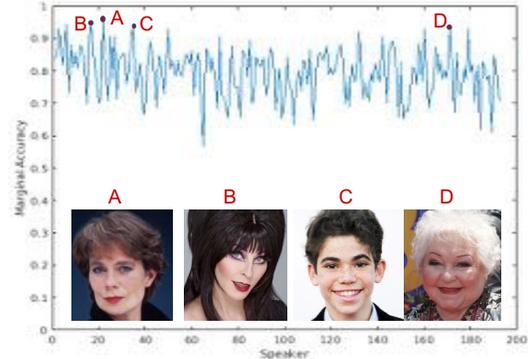

Figure 4: Marginal Accuracies: Marginal accuracies are computed for all speakers in the test set for the static V-F task. The highest marginal accuracies are for A. Celia Imrie, B. Cassandra Peterson, C. Cameron Boyce and D. Estelle Harris.

**Comparison to the Human Benchmark:** The AMT studies show good human performance on the static test set without GNA-variation removal. As can be seen in table 2, the model is comparable to human performance on this task. On the more challenging test set with GNA-variation removed, however, human performance is significantly lower. Interestingly, on this setting the model manages to exceed human performance, which may suggest the presence of *subtle* cross-modal biometrics that are difficult for un-trained humans to identify. We note however, that it is difficult to eliminate all biases that may be exploitable by humans/algorithms performing this task. For instance, since the images in our dataset are sourced from the identities of celebrities (which may occasionally be recognised by the workers), there is likely to be a degree of positive bias in the results from the human study. Moreover, as a consequence of the large-scale nature of the data, it may be possible for the model to learn to use other correlated factors which would be difficult to detect without extensive annotation, e.g. speakers of certain professions, such as sportspeople, may be more likely to be interviewed outside.

**Marginal Accuracies:** An examination of the marginal accuracies of our model shows that some face-voice combinations are significantly more discriminative than others (figure 4). Three speakers who appear to be particularly distinctive, (all with marginal accuracies above 90%), are 'Cameron Boyce', who is a child actor, 'Estelle Harris', who (quoting directly from her Wikipedia page) is 'easily recognised by her distinctive, high-pitched voice' and 'Cassandra Peterson' who portrays the horror hostess character Elvira, Mistress of the Dark. Both Estelle and Cassandra have unconventional hairstyles and make up, as well as highly distinctive manners of speaking. While the task is made easier with particularly distinctive identities such as those mentioned above, we observe that accuracies are above random regardless of the identity of the speaker in the test set, suggesting that the trained model should generalise reasonably well to other speakers.

## 7. Ablation Analysis

**What is the best way to capture articulatory patterns?** We experiment with three different methods of incorporating dynamic features in our architecture: The first computes a single dynamic image per face-track via approximate rank pooling. Since a single face track can contain a long sequence of facial dynamics which can be challenging to capture compactly, we also experiment with the Multiple Dynamic Image (MDI) formulation proposed in [5] in which a sequence of $k$ dynamic images are computed from sets of $m$ contiguous frames at uniformly sampled locations. Each dynamic image is processed independently by the early part of the network and then fused later in the architecture through temporal pooling. In our experiments we take both $k$ and $m$ to be 10, which was found to be most effective in [5]. Differently from the task of action recognition, where the entire video may be needed to inform some actions (e.g. "backhand flick"), we require only local descriptors of motion in order to effectively capture the 'manner of speaking' of a speaker. We therefore also experiment with a third "dense" dynamic image formulation sampled at a fixed stride (thus $m$ remains fixed at 10 but $k$ depends on the length of the face-track).

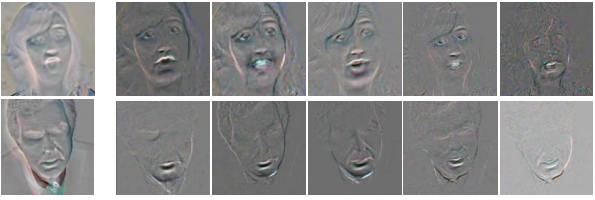

Figure 5: Examples of dynamic images for two identities represented using a single image per face-track (leftmost image in each row) and multiple images per track (5 images on the right). Note how lip motion has clearly been encoded in some of the frames.

| Dynamic Formulation | $A_I$(Total) |
|---|---|
| 1. RGB | $79.2 \pm 0.1$ |
| 2. Dynamic (SDI) | $76.9 \pm 0.6$ |
| 3. Dynamic (MDI) | $79.9 \pm 0.2$ |
| 4. RGB + SDI Fusion | $82.4 \pm 0.3$ |
| 5. RGB + MDI Fusion | $\mathbf{84.3 \pm 0.2}$ |

Table 3: Results using different dynamic formulations for the dynamic matching F-V task; SDI: Single dynamic Image; MDI: Multiple Dynamic Images; The best performance is achieved using both RGB and MDI fusion.

**Temporal Pooling:** To enable the network to ingest multiple dynamic images (MDI), we adopt a simple temporal pooling scheme: for a sequence of input frames $\mathcal{X} = x_1, \ldots, x_T$, we compute a representation $\phi(\mathcal{X}) = \phi_2 \circ \text{pool}(\phi_1(x_1), \ldots, \phi_1(x_T))$, where $\phi_1$ comprises the set of operations up to `pool5` in the face sub-network, followed by a max-pooling over the time dimension before the fully connected layers $\phi_2$ of the network are applied.

**Experiment 1: Video as a collection of static frames:** Our first dynamic experiment simply treats each video as a bag of independent static frames. Since each face-track is a representation of the person speaking, some limited information, including mouth *shape* and facial contortion during speech, can be learnt from individual frames. The static architecture (Figure 2, left) can be used in this case, on RGB frames extracted from the video, with frames extracted in a dense manner with a stride of 6.

**Experiment 2: SDI (Single Dynamic Image) per face-track:** In this experiment a single dynamic image is computed to represent the entire face-track. Since there is a single image per face input, we can once again make direct use of the static architecture (Figure 2, left).

**Experiment 3: MDI (Multiple Dynamic Images) with temporal pooling:** We also experiment with multiple dynamic images in order to capture local variations in motion. In this experiment 10 uniformly sampled dynamic images are extracted per face-track and the static architecture (figure 2, left) with temporal pooling is used.

**Experiment 4: RGB + SDI fusion:** In this experiment we use the dynamic fusion architecture (figure 2, middle). A single RGB frame and dynamic image per track are fed into the network along with the audio segment.

**Experiment 5: RGB + MDI fusion:** We also use dense sampling in order to obtain multiple aligned RGB and dynamic images from each face-track. These inputs are then fed into the fusion architecture, and results are ensembled at test time (this is done in a similar manner to experiment 1, but over the dynamic images as well).

The results on the dynamic test set are given in table 3. To provide a measure of variance, we sampled $10,000$ triplets from the set of all possible triplets on the test identities ten times (with replacement). We report means and std devs for the % accuracy. As expected, using multiple dynamic images instead of a single dynamic image per track leads to a substantial increase in classification accuracy, possibly due to the ability to capture local variations in motion. In order to determine whether the network is learning the face motion of the speaker (and not simply relying on the RGB frame input) we also trained the network with dynamic images only (experiments 2 & 3). As seen from figure 5, it is harder to discern latent variables like age, gender, ethnicity in these images, while mouth motion is clearly encoded. Using these dynamic images alone, we still achieve an accuracy of 77%, suggesting that the network may be able to exploit dynamic cross-modal biometrics.

## 8. Conclusion

In this paper, we have introduced the novel task of cross-modal matching between faces and voices, and proposed a corresponding CNN architecture to address it. Under a binary forced matching constraint, the model is able to match human performance on easy faces and exceed human performance under the more challenging setting in which the speaker pair possesses the same gender, age and nationality. The results of the experiments strongly suggest the existence of *cross-modal* biometric information, leading to the conclusion that perhaps our faces are more similar to our voices than we think.

**Acknowledgements:** The authors gratefully acknowledge the support of EPSRC CDT AIMS EP/L015897/1 and the Programme Grant Seebibyte EP/M013774/1. The authors would also like to thank Erika Lu for help with the AMT study, Hakan Bilen and Joe Levy for useful discussions, and Joon Son Chung for being a living legend.

# References

[1] S. Antol, A. Agrawal, J. Lu, M. Mitchell, D. Batra, C. Zitnick, and D. Parikh. VQA: Visual Question Answering. In *Proc. ICCV*, 2015.

[2] R. Arandjelović and A. Zisserman. Look, listen and learn. In *Proc. ICCV*, 2017.

[3] Y. Aytar, L. Castrejon, C. Vondrick, H. Pirsiavash, and A. Torralba. Cross-modal scene networks. *IEEE PAMI*, 2017.

[4] Y. Aytar, C. Vondrick, and A. Torralba. Soundnet: Learning sound representations from unlabeled video. In *Advances in Neural Information Processing Systems*, pages 892–900, 2016.

[5] H. Bilen, B. Fernando, E. Gavves, and A. Vedaldi. Action recognition with dynamic image networks. *(TPAMI)*, 2017.

[6] H. Bilen, B. Fernando, E. Gavves, A. Vedaldi, and S. Gould. Dynamic image networks for action recognition. In *Proc. CVPR*, pages 3034–3042, 2016.

[7] R. Brunelli and D. Falavigna. Person identification using multiple cues. *IEEE transactions on pattern analysis and machine intelligence*, 17(10):955–966, 1995.

[8] H. E. Çetingúl, E. Erzin, Y. Yemez, and A. M. Tekalp. Multimodal speaker/speech recognition using lip motion, lip texture and audio. *Signal processing*, 86(12):3549–3558, 2006.

[9] H. E. Cetingul, Y. Yemez, E. Erzin, and A. M. Tekalp. Robust lip-motion features for speaker identification. In *Proc. ICASSP*, volume 1, pages I–509. IEEE, 2005.

[10] K. Chatfield, V. Lempitsky, A. Vedaldi, and A. Zisserman. The devil is in the details: an evaluation of recent feature encoding methods. In *Proc. BMVC.*, 2011.

[11] J. S. Chung. Out of time: automated lip sync in the wild. In *Workshop on Multi-view Lip-reading, ACCV*, 2016.

[12] J. S. Chung, A. Senior, O. Vinyals, and A. Zisserman. Lip reading sentences in the wild. In *Proc. CVPR*, 2017.

[13] E. Cvejic, J. Kim, and C. Davis. Recognizing prosody across modalities, face areas and speakers: Examining perceivers sensitivity to variable realizations of visual prosody. *Cognition*, 122(3):442–453, 2012.

[14] N. Dehak, P. J. Kenny, R. Dehak, P. Dumouchel, and P. Ouellet. Front-end factor analysis for speaker verification. *IEEE Transactions on Audio, Speech, and Language Processing*, 19(4):788–798, 2011.

[15] B. Fernando, H. Bilen, E. Gavves, and S. Gould. Self-supervised video representation learning with odd-one-out networks. *arXiv preprint arXiv:1611.06646*, 2016.

[16] A. Frome, G. S. Corrado, J. Shlens, S. Bengio, J. Dean, T. Mikolov, et al. Devise: A deep visual-semantic embedding model. In *Advances in neural information processing systems*, pages 2121–2129, 2013.

[17] H. Hollien and G. P. Moore. Measurements of the vocal folds during changes in pitch. *Journal of Speech, Language, and Hearing Research*, 3(2):157–165, 1960.

[18] S. Ji, W. Xu, M. Yang, and K. Yu. 3d convolutional neural networks for human action recognition. *PAMI*, 35(1):221–231, 2013.

[19] M. Kamachi, H. Hill, K. Lander, and E. Vatikiotis-Bateson. Putting the face to the voice': Matching identity across modality. *Current Biology*, 13(19):1709–1714, 2003.

[20] A. Karpathy and L. Fei-Fei. Deep visual-semantic alignments for generating image descriptions. In *Proc. CVPR*, pages 3128–3137, 2015.

[21] I. Kemelmacher-Shlizerman, S. M. Seitz, D. Miller, and E. Brossard. The megaface benchmark: 1 million faces for recognition at scale. In *Proc. CVPR*, pages 4873–4882, 2016.

[22] E. Khoury, L. El Shafey, C. McCool, M. Günther, and S. Marcel. Bi-modal biometric authentication on mobile phones in challenging conditions. *Image and Vision Computing*, 32(12):1147–1160, 2014.

[23] R. Kiros, R. Salakhutdinov, and R. Zemel. Unifying visual-semantic embeddings with multimodal neural language models. *arXiv preprint arXiv:1411.2539*, 2014.

[24] R. M. Krauss, R. Freyberg, and E. Morsella. Inferring speakers physical attributes from their voices. *Journal of Experimental Social Psychology*, 38(6):618–625, 2002.

[25] G. Kulkarni, V. Premraj, V. Ordonez, S. Dhar, S. Li, Y. Choi, A. C. Berg, and T. Berg. Babytalk: Understanding and generating simple image descriptions. *IEEE PAMI*, 35(12):2891–2903, 2013.

[26] L. Lachs and D. B. Pisoni. Specification of cross-modal source information in isolated kinematic displays of speech. *The Journal of the Acoustical Society of America*, 116(1):507–518, 2004.

[27] K. Lander, H. Hill, M. Kamachi, and E. Vatikiotis-Bateson. It's not what you say but the way you say it: matching faces and voices. *Journal of Experimental Psychology: Human Perception and Performance*, 33(4):905, 2007.

[28] N. Le and J.-M. Odobez. Improving speaker turn embedding by crossmodal transfer learning from face embedding. *arXiv preprint arXiv:1707.02749*, 2017.

[29] X. Lin and D. Parikh. Don't just listen, use your imagination: Leveraging visual common sense for non-visual tasks. In *Proc. CVPR*, pages 2984–2993, 2015.

[30] A. Mahendran and A. Vedaldi. Understanding deep image representations by inverting them. In *Proceedings of the IEEE conference on computer vision and pattern recognition*, pages 5188–5196, 2015.

[31] M. Malinowski and M. Fritz. A multi-world approach to question answering about real-world scenes based on uncertain input. In *NIPS*, pages 1682–1690, 2014.

[32] S. Mallat. Understanding deep convolutional networks. *Phil. Trans. R. Soc. A*, 374(2065):20150203, 2016.

[33] A. Nagrani, J. S. Chung, and A. Zisserman. Voxceleb: a large-scale speaker identification dataset. In *INTERSPEECH*, 2017.

[34] H. Ouyang and T. Lee. A new lip feature representation method for video-based bimodal authentication. In *Proceedings of the 2005 NICTA-HCSNet Multimodal User Interaction Workshop-Volume 57*, pages 33–37. Australian Computer Society, Inc., 2006.

[35] A. Owens, W. Jiajun, J. McDermott, W. Freeman, and A. Torralba. Ambient sound provides supervision for visual learning. In *Proc. ECCV*, 2016.

[36] O. M. Parkhi, A. Vedaldi, and A. Zisserman. Deep face recognition. In *Proc. BMVC.*, 2015.

[37] L. D. Rosenblum, N. M. Smith, S. M. Nichols, S. Hale, and J. Lee. Hearing a face: Cross-modal speaker matching using isolated visible speech. *Perception & psychophysics*,

68(1):84–93, 2006.

[38] A. Roy and S. Marcel. Introducing crossmodal biometrics: Person identification from distinct audio & visual streams. In *Biometrics: Theory Applications and Systems (BTAS), 2010 Fourth IEEE International Conference on*, pages 1–6. IEEE, 2010.

[39] G. Saon, H. Soltau, D. Nahamoo, and M. Picheny. Speaker adaptation of neural network acoustic models using i-vectors. In *ASRU*, pages 55–59, 2013.

[40] S. M. Sheffert and E. Olson. Audiovisual speech facilitates voice learning. *Attention, Perception, & Psychophysics*, 66(2):352–362, 2004.

[41] K. Simonyan and A. Zisserman. Two-stream convolutional networks for action recognition in videos. In *NIPS*, pages 568–576, 2014.

[42] K. Simonyan and A. Zisserman. Very deep convolutional networks for large-scale image recognition. *arXiv preprint arXiv:1409.1556*, 2014.

[43] H. M. Smith, A. K. Dunn, T. Baguley, and P. C. Stacey. Concordant cues in faces and voices: Testing the backup signal hypothesis. *Evolutionary Psychology*, 14(1):1474704916630317, 2016.

[44] H. M. Smith, A. K. Dunn, T. Baguley, and P. C. Stacey. Matching novel face and voice identity using static and dynamic facial images. *Attention, Perception, & Psychophysics*, 78(3):868–879, 2016.

[45] D. Snyder, D. Garcia-Romero, D. Povey, and S. Khudanpur. Deep neural network embeddings for text-independent speaker verification. *Proc. Interspeech 2017*, pages 999–1003, 2017.

[46] Y. Sun, Y. Chen, X. Wang, and X. Tang. Deep learning face representation by joint identification-verification. In *NIPS*, pages 1988–1996, 2014.

[47] Y. Taigman, M. Yang, M. Ranzato, and L. Wolf. Deep-Face: Closing the gap to human-level performance in face verification. In *IEEE CVPR*, 2014.

[48] R. Thornhill and A. P. Møller. Developmental stability, disease and medicine. *Biological Reviews*, 72(4):497–548, 1997.

[49] S. Venugopalan, H. Xu, J. Donahue, M. Rohrbach, R. Mooney, and K. Saenko. Translating videos to natural language using deep recurrent neural networks. *arXiv preprint arXiv:1412.4729*, 2014.

[50] O. Vinyals, A. Toshev, S. Bengio, and D. Erhan. Show and tell: A neural image caption generator. In *CVPR*, pages 3156–3164, 2015.

[51] J. Wang, K. Markert, and M. Everingham. Learning models for object recognition from natural language descriptions. In *Proc. BMVC.*, 2009.

[52] T. Wells, T. Baguley, M. Sergeant, and A. Dunn. Perceptions of human attractiveness comprising face and voice cues. *Archives of sexual behavior*, 42(5):805–811, 2013.

[53] H. Yehia, P. Rubin, and E. Vatikiotis-Bateson. Quantitative association of vocal-tract and facial behavior. *Speech Communication*, 26(1):23–43, 1998.

[54] M. D. Zeiler and R. Fergus. Visualizing and understanding convolutional networks. In *European conference on computer vision*, pages 818–833. Springer, 2014.

[55] J. Zhang, Z. Lin, J. Brandt, X. Shen, and S. Sclaroff. Top-down neural attention by excitation backprop. In *European Conference on Computer Vision*, pages 543–559. Springer, 2016.

[56] G. Zhao and M. Pietikäinen. Visual speaker identification with spatiotemporal directional features. In *International Conference Image Analysis and Recognition*, pages 1–10. Springer, 2013.

[57] C. Zitnick, D. Parikh, and L. Vanderwende. Learning the visual interpretation of sentences. In *Proc. ICCV*, pages 1681–1688, 2013.

## A. Further Ablation Experiments

**What is the minimum duration of audio segment for reliable performance?**

We ran experiments using the best static model with audio segments of $2/3, 1, 2, 3$ and $4$ seconds extracted at both train and test time[4] (results are reported in table 4). We were unable to achieve convergence with $2/3$s, suggesting the minimum duration to be between $2/3$s and $1$s. For longer durations, we found our method to be relatively robust to variations in length.

| Duration (s) | 2/3 | 1 | 2 | 3 | 4 |
|---|---|---|---|---|---|
| Accuracy (%) | 52.8 (chance= 50) | 77.4 | 78.6 | 79.2 | 79.1 |

Table 4: Results on the static matching task using different audio segment lengths.

**Which factor (gender, age, nationality) has the highest effect on performance?**

We conducted an experiment in which test triplets are matched on each factor separately, e.g. for age (A), all elements of a triplet have a similar age. We report the following numbers (% acc.) for the best static model: original performance $81.0$, nationality (N) $78.2$, age (A) $75.0$, gender (G) $65.2$, matching all three (GNA) $63.9$. Matching on gender has the greatest effect on performance. This could be because (i) the dataset [33] does not have enough age and nationality variation (as demonstrated in appendix B); and (ii) gender is more discriminative a factor for the task (especially given that nationality is not always indicative of accent). This is also supported by the visual results in figure 9, which show that the most highly ranked pairs classified correctly are those of different genders.

## B. Test set statistics

As described in section 4 of the paper, the nationalities of the speakers in the test set were obtained by crawling Wikipedia. Figure 6 illustrates the distribution of nationality and gender across this data. In order to create the more challenging *GNA-var removed* dataset, we use only US nationals (of both genders) between the ages of 30-50 years old. The age estimates were obtained manually. The final *GNA-var removed* dataset has 110 speakers.

## C. Amazon Mechanical Turk Study

In this section, we describe in more detail the experimental methodology employed to establish a benchmark for human performance on the test set, described in section 5.2. A pool of Amazon Mechanical Turk (AMT) workers were presented with a set of matching problems, each consisting of two face images and a single audio segment. In each matching problem, they were asked to choose which of the faces corresponded to the person speaking. To ensure that workers were actually listening to the audio segments, a result could only be submitted once the audio sample had been played (achieved by deactivating the selection buttons). Workers could listen to the audio samples as many times as required.

---
[4] To investigate lengths up to 4s and perform a fair comparison, we restricted the dataset to speech segments that were 5s or longer (74% of the total dataset). This ensures the size of the dataset is fixed for each experiment. At test time, sub-segments are densely sampled with the given duration and predictions are averaged.

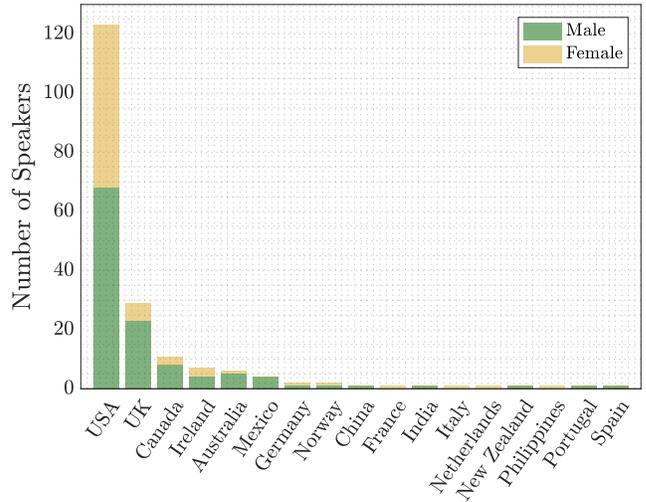

Figure 6: Distribution of nationalities of the speakers in the test set.

In total 500 test triplets were selected, and shown to twenty workers in batches of five (in order to prevent worker fatigue). To avoid the workers 'learning' the face-voice pairings, batches were chosen to ensure the same speaker was not present in the same batch. The accuracy was then computed for each worker over all the triplets that they labelled, and averaged across all workers to produce an estimate of human accuracy. If a worker achieved an accuracy below $40\%$, their results were discarded. To obtain a measure of variance, the mean standard deviation of worker accuracy on each test triplet was calculated and was found to be $2.55\%$. A screenshot of the webpage seen by workers is shown in figure 7.

## D. Network Architecture Details

The filter and output sizes for both voice and face sub-networks can be seen in figure 8.

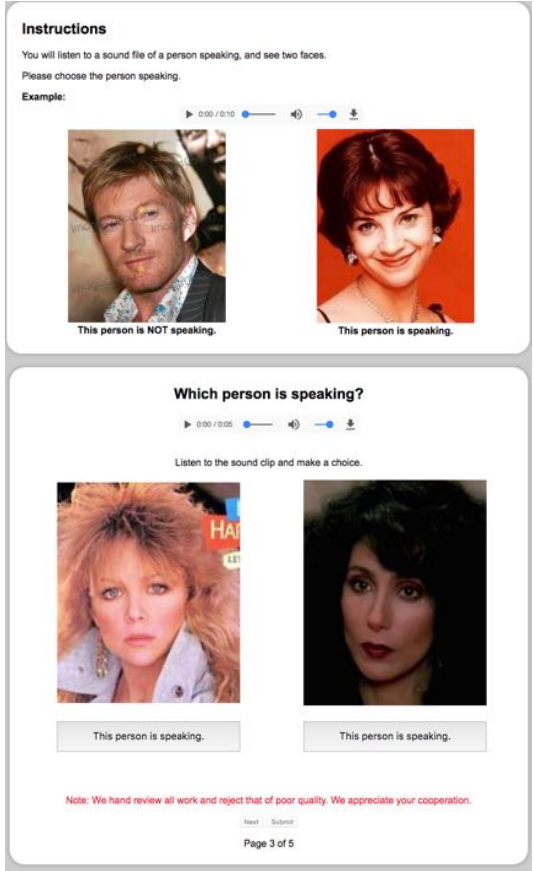

Figure 7: **Screenshot of the webpage shown to workers**. (Top) Upon starting each batch, each worker was provided with instructions and an illustrative example. (Bottom) The interface for submitting worker predictions.

## E. Salient regions

Given the strong performance of the static image model on the challenging *GNA-var removed* evaluation set (as discussed in section 6 of the paper), we would like to gain some insight into how the network is accomplishing the task. The interpretation of the model class of deep neural networks remains a challenging topic and an area of active research (see e.g. [30, 32, 54]). One approach to understanding the decision making process of the network is through region saliency, in which the goal is to identify regions in input space which have exerted maximal influence on a classification decision. Here, we employ the *Excitation Backprop* method introduced in [55] to find discriminative regions in the face inputs[5]. Specifically, we use the *contrastive attention* technique introduced in [55] to visualise saliency following the `relu3` layer in the face streams (we found that higher layers were less informative, typically produc-ing a response covering the extent of the face). The resulting visualisations are shown for samples from the *GNA-var removed* test set in figure 10. We observe that the model often finds highly localised regions in the lower half of the face particularly salient for voice matching (first two rows). However, we also found that in certain cases, it is strongly influenced by a region of greater spatial extent, including the nose and cheeks (third row), or combinations of distinctive features, such as the eyes and mouth (fourth row). Rather than depending on a single consistent feature to solve the task, it therefore seems that the network has learned to draw selectively from a range of signals to classify voices robustly.

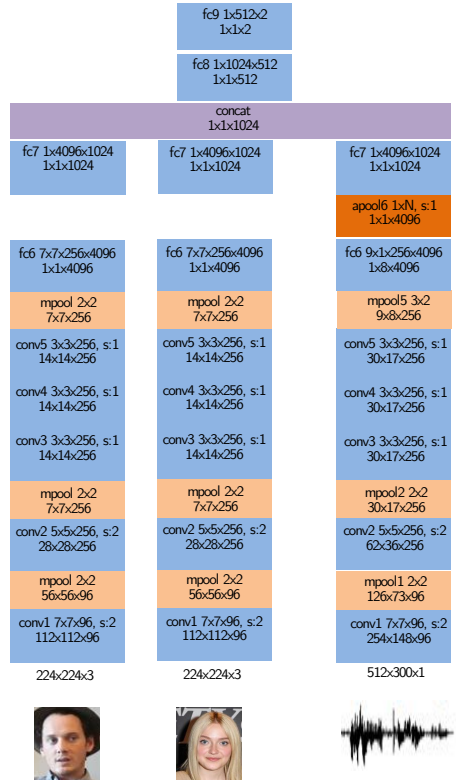

Figure 8: Static architecture for forced matching between two faces and one voice segment (V-F formulation). Note how average pooling is used in the voice subnetwork to deal with variable length speech segments. The value of N in `apool6` changes according to the size of the speech segment input. (Output sizes up till `apool6` are shown for an input speech segment of three seconds, for which $N = 8$ in `apool6`.)

---

[5]Unfortunately voice data, which is consumed by the model in the form of a spectrogram, is less amenable to visual interpretation.

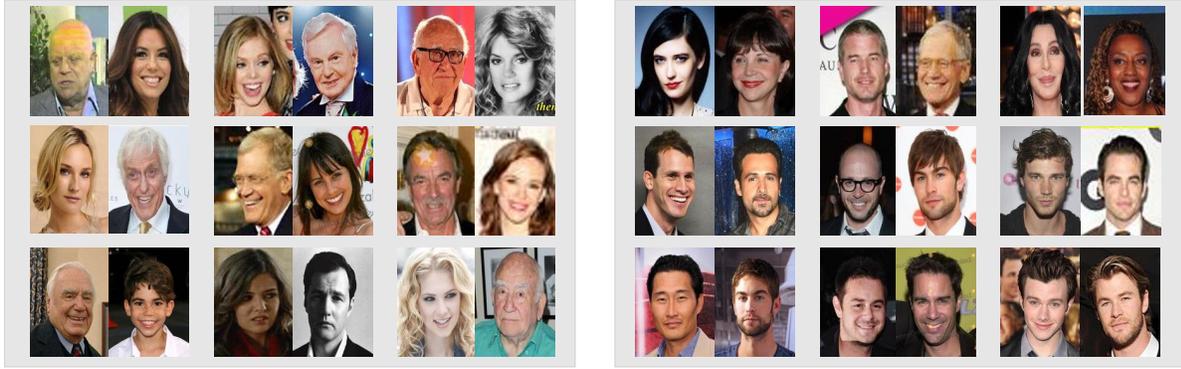

Figure 9: Examples of the top ranked face pairs that were classified correctly using a single voice segment (left panel) and the bottom ranked classified incorrectly (right panel) on the static test set. From the images on the left, it is clear that the model finds it easier to distinguish between faces of different gender and age.

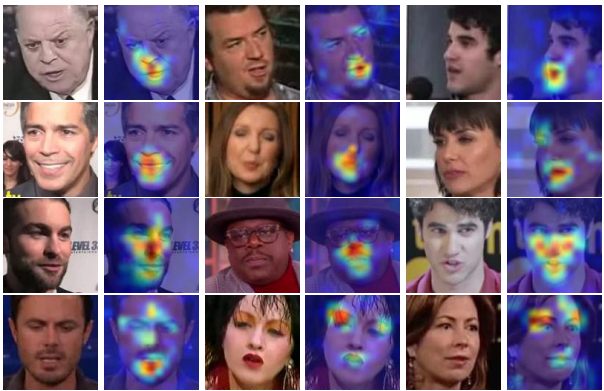

Figure 10: **Salient facial regions for voice classification** - Each example depicts a sampled input face (left) and its corresponding saliency map for the voice matching task (right). The first two rows show highly localised discriminative regions in the lower portion of the face. The final two rows show a more distributed response and usage of other features, particularly eyes. See text for further discussion.